\documentclass[letterpaper]{article} 
\usepackage{aaai25}  
\usepackage{times}  
\usepackage{helvet}  
\usepackage{courier}  
\usepackage[hyphens]{url}  
\usepackage{graphicx} 
\usepackage{natbib}  
\usepackage{caption} 
\newcommand{\adv}{\operatornamewithlimits{adv}}
\newcommand{\trac}{\operatornamewithlimits{trace}}

\newcommand{\normaldist}{\mathcal{N}(\mathbf{0}, \mathbf{I})}
\newcommand{\sign}{\operatornamewithlimits{sign}}
\newcommand{\argmin}{\operatornamewithlimits{argmin}}

\newcommand{\first}[1]{\textbf{#1}}
\newcommand{\second}[1]{\underline{#1}}

\usepackage{amsmath}
\usepackage{amssymb}
\usepackage{amsfonts}

\usepackage{algorithm}
\usepackage{algorithmic}

\usepackage{multirow}
\usepackage{booktabs}
\usepackage{graphicx}
\usepackage{subcaption}

\usepackage{xcolor}

\usepackage{marvosym}

\usepackage{newfloat}
\usepackage{listings}

\DeclareCaptionStyle{ruled}{labelfont=normalfont,labelsep=colon,strut=off} 
\floatstyle{ruled}
\newfloat{listing}{tb}{lst}{}
\floatname{listing}{Listing}
\title{Pixel Is Not a Barrier: An Effective Evasion Attack for Pixel-Domain \\ Diffusion Models}
\author{
    Chun-Yen Shih\textsuperscript{\rm 1, \rm 3}\equalcontrib, 
    Li-Xuan Peng\textsuperscript{\rm 3}\equalcontrib,
    Jia-Wei Liao\textsuperscript{\rm 1, \rm 3},
    Ernie Chu\textsuperscript{\rm 2, \rm 3 \thanks{Work done as research assistant at CITI, Academia Sinica.}}, \\
    Cheng-Fu Chou\textsuperscript{\rm 1},
    Jun-Cheng Chen\textsuperscript{\rm 3 \thanks{Corresponding author. \Letter~pullpull@citi.sinica.edu.tw}
}
}

\affiliations{
    \textsuperscript{\rm 1} National Taiwan University,
    \textsuperscript{\rm 2} Johns Hopkins University, \\
    \textsuperscript{\rm 3} Research Center for Information Technology Innovation, Academia Sinica \\

}

\usepackage{bibentry}

\begin{document}

\maketitle
\begin{abstract}
Diffusion Models have emerged as powerful generative models for high-quality image synthesis, with many subsequent image editing techniques based on them. However, the ease of text-based image editing introduces significant risks, such as malicious editing for scams or intellectual property infringement. Previous works have attempted to safeguard images from diffusion-based editing by adding imperceptible perturbations. These methods are costly and specifically target prevalent Latent Diffusion Models (LDMs), while Pixel-domain Diffusion Models (PDMs) remain largely unexplored and robust against such attacks. Our work addresses this gap by proposing a novel attack framework, AtkPDM. AtkPDM is mainly composed of a feature representation attacking loss that exploits vulnerabilities in denoising UNets and a latent optimization strategy to enhance the naturalness of adversarial images. Extensive experiments demonstrate the effectiveness of our approach in attacking dominant PDM-based editing methods (e.g., SDEdit) while maintaining reasonable fidelity and robustness against common defense methods. Additionally, our framework is extensible to LDMs, achieving comparable performance to existing approaches. Our project page is available at \url{https://alexpeng517.github.io/AtkPDM}.
\end{abstract}

\section{Introduction}

\begin{figure}[t]
    \centering
    \includegraphics[width=\linewidth]{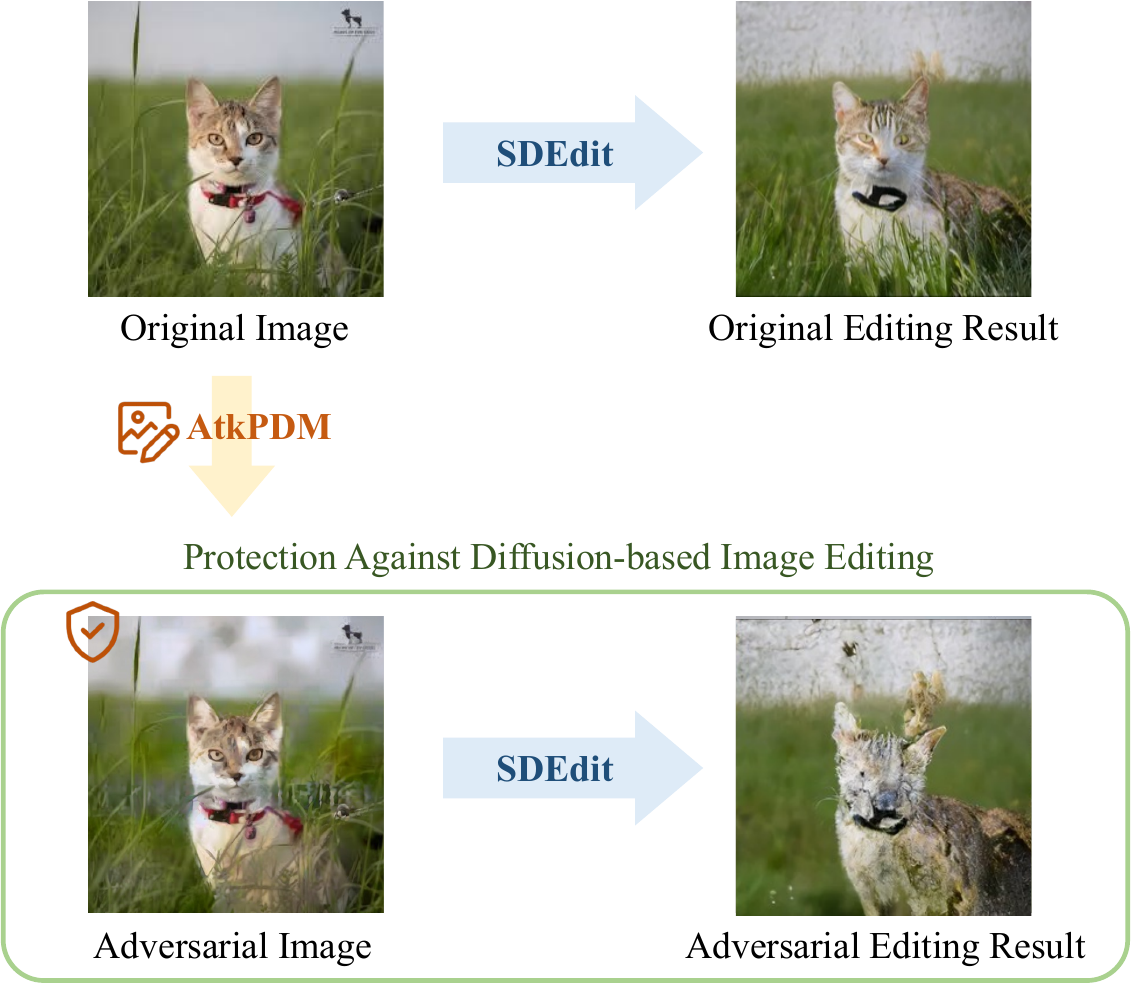}
    \caption{Overview of our attack scenario. Diffusion-based image editing can generate high-quality image variation based on the clean input image. However, by adding carefully crafted perturbation to the clean image, the diffusion process will be disrupted, producing a corrupted image or unrelated image semantics to the original image.}
    \label{fig:teaser}
\end{figure}

In recent years, Generative Diffusion Models (GDMs)~\cite{ho2020denoising, song2021denoisingdiffusionimplicitmodels} emerged as powerful generative models that can produce high-quality images, propelling advancements in image editing and artistic creations. The \textit{ease} of using these models to edit~\cite{meng2021sdedit, wang2023stylediffusion, zhang2023inversion} or synthesize new images~\cite{dhariwal2021diffusion, rombach2022high} has raised concerns about potential malicious usage and intellectual property infringement. For example, malicious users could effortlessly craft fake images with someone's identity or mimic the style of a specific artist. An effective protection against these threats is to craft an adversarial image to force the diffusion model to generate corrupted images or unrelated images to the original inputs. Researchers have made significant strides in crafting imperceptible adversarial perturbations on images to protect against diffusion-based editing. 

Previous works such as PhotoGuard~\cite{salman2023raisingcostmaliciousaipowered} and Glaze~\cite{shan2023glazeprotectingartistsstyle} have effectively attacked Latent Diffusion Models (LDMs) by minimizing the latent distance between the protected images and their target counterparts. PhotoGuard first introduces attacking either encoders or diffusion process in LDMs via Projected Gradient Descent (PGD)~\cite{madry2018towards} for the protection purpose; however, it requires backpropagating the entire diffusion process, making it prohibitively expensive. Subsequent works AdvDM~\cite{liang2023adversarialexampledoesgood} and Mist~\cite{liang2023mistimprovedadversarialexamples} leverage the semantic loss and textural loss combined with Monte Carlo method to craft adversarial images both effectively and efficiently. Diff-Protect ~\cite{xue2024diffusion} further improve adversarial effectiveness and optimization speed via Score Distillation Sampling (SDS) ~\cite{poole2022dreamfusiontextto3dusing2d}, setting the state-of-the-art performance on LDMs.

However, previous works primarily focus on LDMs, and attacks on Pixel-domain Diffusion Models (PDMs) remain unexplored. Xue et al. ~\cite{xue2024diffusion} also highlighted a critical limitation of current methods: the attacking effectiveness is mainly attributed to the vulnerability of the VAE encoders in LDM; however, PDMs don't have such encoders, making current methods hard to transfer to PDMs. The latest work \cite{xue2024pixelbarrierdiffusionmodels} has attempted to attack PDMs, but the result suggests that PDMs are robust to pixel-domain perturbations. Our goal is to mitigate the gap between these limitations.

In this paper, we propose an innovative framework AtkPDM, to effectively attack PDMs. Our approach includes a novel \textbf{feature attacking loss} that exploits the vulnerabilities in denoising UNet to distract the model from recognizing the correct semantics of the image, a \textbf{fidelity loss} that acts as optimization constraints that ensure the imperceptibility of adversarial image and controls the attack budget, and a \textbf{latent optimization strategy} utilizing victim-model-agnostic VAEs to further enhance the naturalness of our adversarial image. With extensive experiments on different PDMs, the results show that our method is effective and affordable while robust to prevalent defense methods and exhibiting attack transferability in the black-box setting. In addition, our approach outperforms current semantic-loss-based and PGD-based methods, reaching state-of-the-art performance on attacking PDMs. Our contributions are summarized as follows:

\begin{enumerate}
    \item We propose a novel attack framework targeting PDMs, achieving state-of-the-art performance in safeguarding images from being edited by SDEdit.
    \item  We propose a novel feature attacking loss design to distract UNet feature representation effectively.
    \item We propose a latent optimization strategy via model-agnostic VAEs to enhance the naturalness of our adversarial images.
\end{enumerate}

\section{Related Work}

\subsection{Image Editing with SDEdit-based Methods}
With the multi-step sampling nature and the ease of converting a sample to intermediate noisy latent via forward diffusion of Diffusion Models~\cite{ho2020denoising}. SDEdit ~\cite{meng2021sdedit} indicates that the diffusion model sampling process is not necessarily required to begin with random Gaussian noise, but allows starting with a mixture of input image and noise at arbitrary strength, i.e. forwarded to $t \in[0, T]$, for the editing. This technique is generalized to both PDMs and LDMs. Subsequent editing frameworks~\cite{hertz2023prompttoprompt, tumanyan2023plug, parmar2023zero, Mokady_2023_CVPR} also build upon this concept.

\subsection{Evasion Attack for Diffusion Model}
To counteract SDEdit-based editing, Salman et al. first proposed PhotoGuard~\cite{salman2023raisingcostmaliciousaipowered} to introduce two attacking paradigms based on Projected Gradient Descent (PGD)~\cite{madry2018towards}. The first is the Encoder Attack, which aims to disrupt the latent representations of the Variational Autoencoder (VAE) of the LDMs, and the second is the Diffusion Attack, which focuses more on disrupting the entire diffusion process of the LDMs. The Encoder Attack is simple yet effective, but the attacking results are sub-optimal due to its less flexibility for optimization than the Diffusion Attack. Although the Diffusion Attack achieves better attack results, it is prohibitively expensive due to its requirement of backpropagation through all the diffusion steps. In the following, we introduce other proposed method targeting different aspects for attacking diffusion models.

\subsubsection{Diffusion Attacks.}
Despite the cost of performing the Diffusion Attack, the higher generalizability and universally applicable nature drive previous works focusing on disrupting the process with lower cost. Liang et al.~\cite{liang2023adversarialexampledoesgood} proposed AdvDM to utilize the diffusion training loss as their attacking semantic loss. Then, AdvDM performs gradient ascent with the Monte Carlo method, aiming to disrupt the denoising process without calculating full backpropagation. Mist~\cite{liang2023mistimprovedadversarialexamples} also incorporates semantic loss and performs constrained optimization via PGD to achieve better attacking performance.

\subsubsection{Encoder Attacks.}
On the other hand, researchers found that VAEs in widely adopted LDMs are more vulnerable to attack at a lower cost than the expensive diffusion process. Hence, they~\cite{salman2023raisingcostmaliciousaipowered, liang2023mistimprovedadversarialexamples, shan2023glazeprotectingartistsstyle, xue2024effectiveprotectiondiffusionbased} focus on disrupting the latent representation in LDM via PGD and highlight the encoder attacks are more effective against LDMs.

\subsubsection{Conditional Module Attacks.}
Most of the LDMs contain conditional modules for steering generation, previous works~\cite{shan2023glazeprotectingartistsstyle, shan2024nightshadepromptspecificpoisoningattacks, lo2024distraction} exploited the vulnerability of text conditioning modules. By disrupting the cross-attention between text concepts and image semantics, these methods effectively interfere with the diffusion model's ability to capture image-text alignment, thereby achieving the attack.

\subsubsection{Limitations of Current Methods.}
To the best of our knowledge, previous works primarily focus on adversarial attacks for LDMs, while attacks on PDMs remain unexplored. Xue et al.~\cite{xue2024pixelbarrierdiffusionmodels} further emphasized the difficulty of attacking PDMs. However, in our work, we find that by crafting an adversarial image to corrupt the intermediate representation of diffusion UNet, we can achieve promising attack performance for PDMs, 
while the attack is also compatible with LDMs. Moreover, inspired by~\cite{laidlaw2021perceptual, liu2023instruct2attack} which utilize LPIPS~\cite{zhang2018unreasonable} as the distortion measure, we also propose a novel attacking loss as the measure to craft better adversarial images for PDMs.

\begin{figure}[t]
    \centering
    \includegraphics[width=1\linewidth]{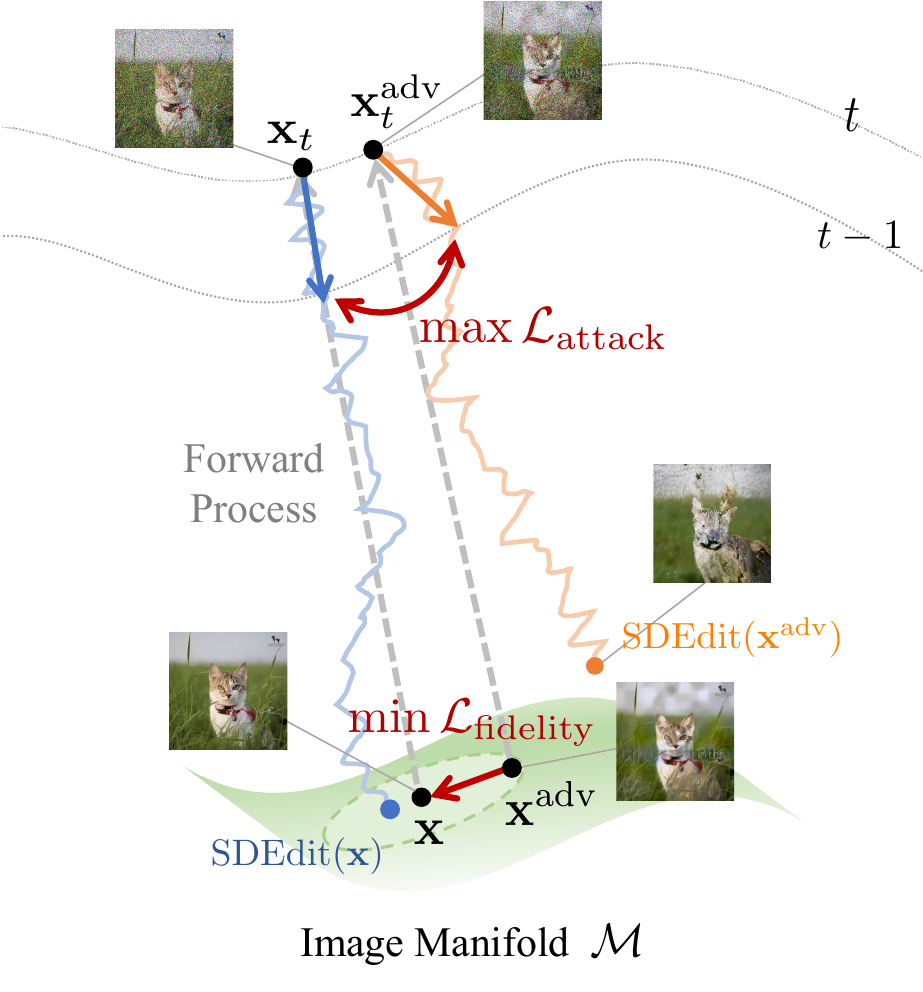}
    \caption{Conceptual illustration of our method. We randomly forward both the clean image $\mathbf{x}$ and adversarial image $\mathbf{x}^{\adv}$ to noise level $t$, then utilize our feature attacking loss to maximize the feature distance between noisy latent $\mathbf{x}_t$ and $\mathbf{x}^{\adv}_t$ in the reverse process of diffusion models while imposing our fidelity loss as a constraint to ensure the adversarial image from being deviated from the original image. We update the $\mathbf{x}^{\adv}$ in latent space instead of in pixel space to ensure the naturalness of $\mathbf{x}^{\adv}$.}
    \label{concept}
\end{figure}

\section{Methodology}

\subsection{Threat Model and Problem Setting}
The malicious user collects an image $\mathbf{x}$ from the internet and uses SDEdit \cite{meng2021sdedit} to generate unauthorized image translations or editing, denoted as $\text{SDEdit}(\mathbf{x}, t)$, that manipulates the original input image $\mathbf{x}$. Our work aims to safeguard the input image $\mathbf{x}$ from the unauthorized manipulations by crafting an adversarial image $\mathbf{x}^{\adv}$ through adding imperceptible perturbation to disrupt the reverse diffusion process of SDEdit for corrupted editions. For example, we want the main object of the image, e.g., the cat in the source image $\mathbf{x}$ as shown in Figure~\ref{concept} is unable to be reconstructed by the reverse diffusion process. Meanwhile, the adversarial image should maintain similarity to the source image to ensure fidelity. The reason why we target SDEdit as our threat model is that it is recognized as the most common and general operation in diffusion-based unconditional image translation and conditional image editing. Additionally, it has been incorporated into various editing pipelines~\cite{tsaban2023leditsrealimageediting, zhang2023inversion}. Here we focus on the unconditional image translation for our main study, as they are essential in both unconditional and conditional editing pipelines. Formally, our objective to effectively safeguard images while maintaining fidelity is formulated as:

\begin{equation}
    \begin{aligned}
        & \max_{\mathbf{x}^{\adv} \in \mathcal{M}} d(\text{SDEdit}(\mathbf{x}, t), \text{SDEdit}(\mathbf{x}^{\adv}, t)) \\
    & \text{subject to } d^{\prime}(\mathbf{x}, \mathbf{x}^{\adv}) \leq \delta,
    \end{aligned}
    \label{eq:original_probelm_setting_constraint}
\end{equation}

where $\mathcal{M}$ indicates natural image manifold, $d$ and $d^{\prime}$ indicate image distance functions, and $\delta$ denotes the fidelity budget.

In the following sections, we first present a conceptual illustration of our method, followed by our framework for solving the optimization problem. We then discuss the novel design of our attacking loss and fidelity constraints, which provide more efficient criteria compared to previous methods. Finally, we introduce an advanced design to enhance adversarial image quality by latent optimization via victim-model-agnostic VAE.

\begin{figure*}[t]
    \centering
    \includegraphics[width=1\linewidth]{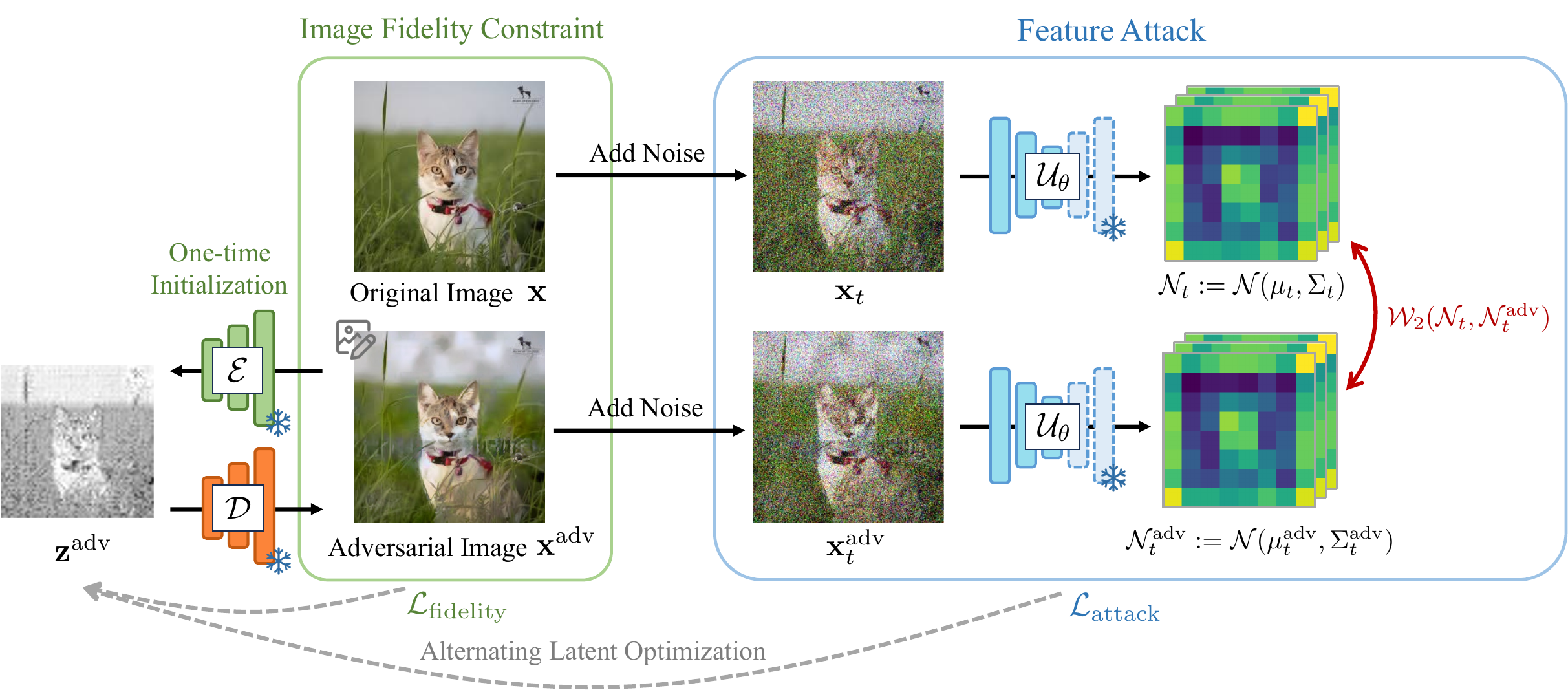}
    \caption{Overview of our AtkPDM$^{+}$ algorithm: Starting from the latent, $\mathbf{z}^{\adv}$, of the initial adversarial image, we first decode back to pixel-domain to perform forward diffusion with both $\mathbf{x}$ and $\mathbf{x}^{\adv}$ and feed them to frozen victim UNet. We then extract the feature representation of the middle block in UNet to calculate our $\mathcal{L}_\text{attack}$, aiming to distract the recognition of image semantics. We also calculate our $\mathcal{L}_\text{fidelity}$ in pixel-domain to constrain the optimization. Finally, the $\mathbf{z}^{\adv}$ is being alternatively updated by loss gradients.}
    \label{framework}
\end{figure*}

\subsection{Overview}
To achieve effective protection against diffusion-based editing, we aim to push the adversarial image away from the original clean image by disrupting the intermediate step in the reverse diffusion process. For practical real-world applications, it's essential to ensure the adversarial image is perceptually similar to the original image. In practice, we uniformly sample the value of the forward diffusion step $t \sim [0, T]$ to generate noisy images and then perform optimization to craft the adversarial image $\mathbf{x}^{\adv}$ via our attacking and fidelity losses, repeating the same process $N$ times or until convergence. Figure \ref{concept} depicts these two push-and-pull criteria during different noise levels, the successful attack is represented in the light orange line where the reverse sample moves far away from the normal edition of the image. More specifically, our method can be formulated as follows:

\begin{equation}
    \begin{aligned}
        & \max_{\mathbf{x}^{\adv} \in \mathcal{M}}
        \mathbb{E}_{
        t,
        \mathbf{x}_t| \mathbf{x}, \mathbf{x}_t^{\adv}| \mathbf{x}}
         \mathcal{L}_\text{attack}(\mathbf{x}_t, \mathbf{x}_t^{\adv}) \\
        & \text{subject to } \mathcal{L}_\text{fidelity}(\mathbf{x}, \mathbf{x}^{\adv}) \leq \delta,
    \end{aligned}
    \label{eq:probelm_setting_constraint}
\end{equation}
where $\delta$ denotes the attacking budget. The details of the attacking loss $\mathcal{L}_\text{attack}$ and the fidelity loss $\mathcal{L}_\text{fidelity}$ will be discussed in the following sections.

\subsubsection{Framework.}
Our framework, shown in Figure~\ref{framework}, utilizes two identical and frozen victim UNets to extract feature representations from clean and adversarial images for our attacking loss calculation and a victim-model-agnostic VAE for the latent optimization strategy.

\subsection{Proposed Losses}
We propose two novel losses as optimization objectives to craft an adversarial example efficiently without running through all the diffusion steps. The attacking loss is designed to distract the feature representation of the denoising UNet; The fidelity loss is a constraint to ensure the adversarial image quality. For notation simplicity, we first define the samples $\mathbf{x}, \mathbf{x}^{\adv}$ in different forwarded steps. Let $\mathcal{F}(\mathbf{x}, t, \epsilon) = \sqrt{\bar{\alpha}_t} \mathbf{x} + \sqrt{1-\bar{\alpha}_t} \epsilon$ be the diffusion forward process. Given timestep $t$ sample from $[0, T]$, noises $\epsilon, \epsilon^{\adv}$ sample from $\normaldist$. We denote $\mathbf{x}_t = \mathcal{F}(\mathbf{x}, t, \epsilon)$, and $\mathbf{x}^{\adv}_t = \mathcal{F}(\mathbf{x}^{\adv}, t, \epsilon^{\adv})$.

\subsubsection{Attacking Loss.}
Our goal is to define effective criteria that could finally distract the reverse denoising process. PhotoGuard~\cite{salman2023raisingcostmaliciousaipowered} proposed to backpropagate through all the steps of the reverse denoising process via PGD. However, this approach is prohibitively expensive, Diff-Protect~\cite{xue2024effectiveprotectiondiffusionbased} proposed to avoid the massive cost by leveraging Score Distillation~\cite{poole2022dreamfusiontextto3dusing2d} in optimization. Nevertheless, Diff-Protect relies heavily on gradients of attacking encoder of an LDM as stated in their results. In PDM, we don't have such an encoder to attack; however, we find that the denoising UNet has a similar structure to encoder-decoder models, and some previous works~\cite{lin2024diffusionmodelperceptualloss, li2023fasterdiffusionrethinkingrole} characterize this property to accelerate and enhance the generation. From our observations of the feature roles in denoising UNets, we hypothesize that distracting specific inherent feature representation in UNet blocks could lead to effectively crafting an adversarial image. In practice, we first extract the feature representations of forwarded images $\mathbf{x}_t$ and $\mathbf{x}^{\adv}_t$ in frozen UNet blocks of timestep $t$. Then, we adopt 2-Wasserstein distance~\cite{arjovsky2017wasserstein} to measure the discrepancy in the UNet feature space. The reason for choosing the 2-Wasserstein distance is that it better captures the distributional discrepancy via Optimal Transport Theory~\cite{chen2018optimal}. Note that we aim to maximize the distance between $\mathbf{x}^{\adv}_t$ and $\mathbf{x}_t$ in the UNet feature space to distract the denoising process. Formally, the attacking loss $\mathcal{L}_\text{attack}$ is defined as:

\begin{equation}
    \mathcal{L}_\text{attack}(\mathbf{x}_t, \mathbf{x}^{\adv}_t)
    =\mathcal{W}_2 \left(
    \mathcal{U}^\text{(mid)}_{\theta}(\mathbf{x}_t), \mathcal{U}^\text{(mid)}_{\theta}(\mathbf{x}^{\adv}_t)
    \right).
\end{equation}

Assuming the feature distributions approximate normal distributions expressed by mean $\mu_t$ and $\mu_t^{\adv}$, and non-singular covariance matrices $\Sigma_t$ and $\Sigma_t^{\adv}$. The calculation of the 2-Wasserstein distance between two normal distributions is viable through the closed-form solution~\cite{dowson1982frechet, olkin1982distance, chen2018optimal}:

\begin{equation}
    \begin{aligned}
        & \mathcal{W}_2^2(\mathcal{N}(\mu_t, \Sigma_t), \mathcal{N}(\mu_t^{\adv}, \Sigma_t^{\adv}))
        = \|\mu_t-\mu_t^{\adv}\|_2^2 \\
        & \qquad + \text{trace} (\Sigma_t + \Sigma_t^{\adv}
        -2({\Sigma_t^{\adv}}^{\frac{1}{2}}\Sigma_t{\Sigma_t^{\adv}}^{\frac{1}{2}})^\frac{1}{2} ).
    \end{aligned}
    \label{eq:wasserstein_distance}
\end{equation}

\subsubsection{Fidelity Loss.}
To control the attack budget for adversarial image quality, we design a constraint function that utilizes the feature extractor from a pretrained classifier to calculate the fidelity loss. In our case, we sum up the 2-Wasserstein feature losses of $L$ different layers. Specifically, we define $\mathcal{L}_\text{fidelity}$ as:
\begin{equation}
    \mathcal{L}_\text{fidelity}(\mathbf{x}_t, \mathbf{x}^{\adv}_t)
    = \sum_{\ell=1}^L \mathcal{W}_2(\phi_\ell(\mathbf{x}), \phi_\ell(\mathbf{x}^{\adv})),
\end{equation}
where $\mathcal{W}_2$ denotes 2-Wasserstein distance and $\phi_\ell$ denotes layer $\ell$ of the feature extractor.

\subsection{Alternating Optimization for Adversarial Image}
We solve the constrained optimization problem via alternating optimization to craft the adversarial images, detailed optimization loop of AtkPDM$^{+}$ is provided in Algorithm ~\ref{alg:attdpmplus}. To maximize the $\mathcal{L}_\text{attack}$, we take the negative $\mathcal{L}_\text{attack}$ and perform gradient descent. AtkPDM algorithm and the derivation of the alternating optimization are provided in Appendix.

\subsection{Latent Optimization via Pretrained-VAE}
Previous works suggest that diffusion models have a strong capability against adversarial perturbations~\cite{xue2024pixelbarrierdiffusionmodels}, making them hard to be attacked via pixel-domain optimization. Moreover, they are even considered as good purifiers of adversarial perturbations~\cite{nie2022diffusionmodelsadversarialpurification}.

Here, we propose a strategy that crafts the perturbation in the latent space of the pre-trained Variational Autoencoder (VAE) ~\cite{kingma2014autoencoding}, and the gradients are used to update the latent. After $N$ iterations or losses converge, we decode back via the decoder $\mathcal{D}$ to pixel domain as our final adversarial image. The motivation for adopting VAE is inspired by MPGD~\cite{he2024manifold}. This strategy is effective for crafting a robust adversarial image against pixel-domain diffusion models while also better preserving the adversarial image quality rather than only incorporating fidelity constraints. Note that, ideally, manifold preservation is guaranteed when using perfect VAE. In practice, we use the best available LDM's VAE agnostic to the victim model as our latent optimization VAE. Detailed latent optimization loop is provided in Algorithm~\ref{alg:attdpmplus}.

\begin{algorithm}[t]
    \caption{AtkPDM$^{+}$}
    \label{alg:attdpmplus}
    \small{
    \begin{algorithmic}[1] 
        \STATE{\textbf{Input:}
        Image to be protected $\mathbf{x}$, attack budget $\delta > 0$, step size $\gamma_\text{attack}, \gamma_\text{fidelity}>0$, \textcolor{black}{VAE encoder $\mathcal{E}$, and VAE decoder $\mathcal{D}$}}
        \STATE{\textbf{Initialization:} $\mathbf{x}^{\adv} \leftarrow \mathbf{x}$, $L_\text{attack} \leftarrow \infty$}
        \STATE{\textcolor{black}{Encode adversarial image to latent space: $\mathbf{z}^{\adv} \leftarrow \mathcal{E}(\mathbf{x}^{\adv})$}}
        \WHILE{$L_\text{attack}$ not convergent}
            \STATE{\textcolor{black}{Decode adversarial latent to pixel space: $\mathbf{x}^{\adv} \leftarrow \mathcal{D}(\mathbf{z}^{\adv})$}}
            \STATE{Sample timestep: $t \sim [0, T]$}
            \STATE{Sample noise: $\epsilon, \epsilon^{\adv} \sim \normaldist$}
            \STATE{Compute original noisy sample: \\
            $\mathbf{x}_t \leftarrow \mathcal{F}(\mathbf{x}, t, \epsilon)$}
            \STATE{Compute adversarial noisy sample: \\
            $\mathbf{x}^{\adv}_t \leftarrow \mathcal{F}(\mathbf{x}^{\adv}, t, \epsilon^{\adv})$}
            \STATE{\textcolor{black}{Update $\mathbf{z}^{\adv}$ by Gradient Descent: \\
            $\mathbf{z}^{\adv} \leftarrow \mathbf{z}^{\adv} -
            \gamma_\text{attack} \sign( \nabla_{\mathbf{z}^{\adv}} (-\mathcal{L}_\text{attack}(\mathbf{x}_t, \mathbf{x}^{\adv}_t)))$}}
            \WHILE{$\mathcal{L}_\text{fidelity}(\mathbf{x}, \textcolor{black}{\mathcal{D}(\mathbf{z}^{\adv})}) > \delta$}
            \STATE{ 
            \textcolor{black}{$\mathbf{z}^{\adv} \leftarrow \mathbf{z}^{\adv} -
            \gamma_\text{fidelity} \nabla_{\mathbf{z}^{\adv}} \mathcal{L}_\text{fidelity}(\mathbf{x}, \mathcal{D}(\mathbf{z}^{\adv}))$}}
            \ENDWHILE
        \ENDWHILE
        \STATE{\textcolor{black}{Decode adversarial latent to pixel space: $\mathbf{x}^{\adv} \leftarrow \mathcal{D}(\mathbf{z}^{\adv})$}}
        \RETURN {$\mathbf{x}^{\adv}$}
    \end{algorithmic}
    }
\end{algorithm}

\section{Experiment Results}

\begin{table*}[t]
    \centering
    \small{
    \begin{tabular}{ll|ccc|cccc}
        \toprule
        & \multirow{2}{*}{Methods} & \multicolumn{3}{c|}{Adversarial Image Quality} & \multicolumn{4}{c}{Attacking Effectiveness} \\ 
         & & SSIM $\uparrow$ & PSNR $\uparrow$ & LPIPS $\downarrow$ & SSIM $\downarrow$ & PSNR $\downarrow$ & LPIPS $\uparrow$ & IA $\downarrow$ \\
        \midrule
        \multirow{4}{*}{\rotatebox{90}{Church}}
        & AdvDM
        & 0.37 $\pm$ 0.09 & 28.17 $\pm$ 0.22 & 0.73 $\pm$ 0.16 & 0.89 $\pm$ 0.05 & 31.06 $\pm$ 1.94 & 0.17 $\pm$ 0.09 & 0.93 $\pm$ 0.04 \\
        & Diff-Protect
        & 0.39 $\pm$ 0.07 & 28.03 $\pm$ 0.12 & 0.67 $\pm$ 0.11 & 0.82 $\pm$ 0.05 & 31.90 $\pm$ 1.08 & 0.23 $\pm$ 0.07 & 0.91 $\pm$ 0.04 \\
        & AtkPDM (Ours) & \second{0.75} $\pm$ 0.03 & \second{28.22} $\pm$ 0.10 & \second{0.26} $\pm$ 0.04 & \first{0.75} $\pm$ 0.04 & \first{29.61} $\pm$ 0.23 & \first{0.40} $\pm$ 0.05 & \first{0.76} $\pm$ 0.06 \\
        & AtkPDM$^+$ (Ours) & \first{0.81} $\pm$ 0.03 & \first{28.64} $\pm$ 0.19 & \first{0.13} $\pm$ 0.02 & \second{0.79} $\pm$ 0.04 & \second{30.05} $\pm$ 0.47 & \second{0.33} $\pm$ 0.07 & \second{0.81} $\pm$ 0.06 \\
        \midrule
        \multirow{4}{*}{\rotatebox{90}{Cat}}
        & AdvDM
        & 0.48 $\pm$ 0.09 & 28.34 $\pm$ 0.18 & 0.65 $\pm$ 0.12 & 0.96 $\pm$ 0.02 & \second{32.32} $\pm$ 2.49 & 0.10 $\pm$ 0.05 & 0.97 $\pm$ 0.03 \\
        & Diff-Protect
        & 0.33 $\pm$ 0.10 & 28.03 $\pm$ 0.15 & 0.80 $\pm$ 0.15 & \second{0.90} $\pm$ 0.05 & 33.94 $\pm$ 1.93 & \second{0.18} $\pm$ 0.08 & 0.95 $\pm$ 0.03 \\
        & AtkPDM (Ours) & \second{0.71} $\pm$ 0.06 & \second{28.47} $\pm$ 0.18 & \second{0.29} $\pm$ 0.05 & \first{0.83} $\pm$ 0.03 & \first{30.73} $\pm$ 0.51 & \first{0.39} $\pm$ 0.05 & \first{0.81} $\pm$ 0.04 \\
        & AtkPDM$^+$ (Ours) & \first{0.83} $\pm$ 0.04 & \first{29.41} $\pm$ 0.37 & \first{0.09} $\pm$ 0.02 & 0.93 $\pm$ 0.01 & 33.02 $\pm$ 0.74 & \second{0.18} $\pm$ 0.02 & \second{0.92} $\pm$ 0.01\\
        \midrule
        \multirow{4}{*}{\rotatebox{90}{Face}}
        & AdvDM
        & 0.48 $\pm$ 0.05 & \first{28.75} $\pm$ 0.18 & 0.64 $\pm$ 0.10 & 0.99 $\pm$ 0.00 & 37.96 $\pm$ 1.75 & 0.02 $\pm$ 0.01 & 0.99 $\pm$ 0.00 \\
        & Diff-Protect
        & 0.25 $\pm$ 0.04 & 28.09 $\pm$ 0.20 & 0.91 $\pm$ 0.11 & 0.95 $\pm$ 0.02 & 35.33 $\pm$ 1.62 & 0.08 $\pm$ 0.04 & 0.96 $\pm$ 0.02 \\
        & AtkPDM (Ours) & \second{0.56} $\pm$ 0.04 & 28.01 $\pm$ 0.22 & \second{0.36} $\pm$ 0.04 & \first{0.74} $\pm$ 0.03 & \first{29.14} $\pm$ 0.36 & \first{0.40} $\pm$ 0.05 & \first{0.62} $\pm$ 0.07 \\
        & AtkPDM$^+$ (Ours) & \first{0.81} $\pm$ 0.04 & \second{28.39} $\pm$ 0.20 & \first{0.12} $\pm$ 0.03 & \second{0.86} $\pm$ 0.03 & \second{30.26} $\pm$ 0.72 & \second{0.24} $\pm$ 0.07 & \second{0.80} $\pm$ 0.08 \\
        \bottomrule
    \end{tabular}
    }
    \caption{Quantitative results in attacking different unconditional PDMs. The best is marked in bold and the second best is underlined. Errors denote one standard deviation of all images in our test datasets.}
    \label{tab:attackPDM}
\end{table*}

\begin{table*}[t]
    \centering
    \small{
    \begin{tabular}{ll|ccc|cccc}
        \toprule
        & \multirow{2}{*}{Methods} & \multicolumn{3}{c|}{Adversarial Image Quality} & \multicolumn{4}{c}{Attacking Effectiveness} \\ 
         & & SSIM $\uparrow$ & PSNR $\uparrow$ & LPIPS $\downarrow$ & SSIM $\downarrow$ & PSNR $\downarrow$ & LPIPS $\uparrow$ & IA $\downarrow$ \\
        \midrule
        \multirow{4}{*}{\rotatebox{90}{}}
        & Diff-Protect
        & 0.47 $\pm$ 0.08  & 27.96 $\pm$ 0.08  & 0.46 $\pm$ 0.05  & \first{0.49} $\pm$ 0.10 & \first{28.13} $\pm$ 0.15 & \first{0.36} $\pm$ 0.10  & \first{0.79} $\pm$ 0.06 \\
        & AtkPDM$^+$ (Ours) &  \first{0.79} $\pm$ 0.06  & \first{28.48} $\pm$ 0.33  & \first{0.06} $\pm$ 0.02 & 0.72 $\pm$ 0.10  & 28.50 $\pm$ 0.48  & 0.10 $\pm$ 0.04  & 0.86 $\pm$ 0.08 \\
        \bottomrule
    \end{tabular}
    }
    \caption{Quantitative results in attacking conditional PDM DeepFloyd IF. The best is marked in bold and the second best is underlined. Errors denote one standard deviation of all images in our test datasets.}
    \label{tab:deepfloydif}
\end{table*}

\begin{table}[t]
\footnotesize{
    \centering
    \begin{tabular}{lcccc}
        \toprule
        \multirow{2}{*}{Defense Method} & \multicolumn{4}{c}{Attacking Effectiveness} \\ 
         &  SSIM $\downarrow$ & PSNR $\downarrow$ & LPIPS $\uparrow$ & IA $\downarrow$ \\
        \midrule
        LDM-Pure & 0.78 & 29.84 & 0.35 & 0.80 \\
        Crop-and-Resize & 0.68 & 29.28 & 0.42 & 0.79 \\
        JPEG Comp. & 0.78  & 29.82 & 0.36 & 0.79 \\
        \midrule
        None & 0.79 & 30.05 & 0.33 & 0.81 \\
        \bottomrule
    \end{tabular}
    \caption{Quantitative results of our adversarial images against defense methods. LDM-Pure, Crop-and-Resize, and JPEG Compression fail to defend our attack. ``None'' indicates no defense is applied, as the baseline for comparison.}
\label{tab:defense}
}
\end{table}

\subsection{Experiment Settings}

\subsubsection{Implementation Details.}
We conduct all our experiments in white box settings and examine the effectiveness of our attacks using SDEdit~\cite{meng2021sdedit}. For the VAE~\cite{kingma2014autoencoding} in our AtkPDM$^{+}$, we utilize the one provided by StableDiffusion v1.5~\cite{rombach2022high}.
We run all of our experiments with 300 optimization steps, which empirically determined, balancing attacking effectiveness and adversarial image quality with a reasonable speed. Other loss parameters and running time are provided in the Appendix. The implementation is built on the Diffusers library~\cite{von-platen-etal-2022-diffusers}. All the experiments are conducted with a single Nvidia Tesla V100 GPU.

\subsubsection{Victim Models and Datasets.}
We test our approach on PDMs with three open-source checkpoints on HuggingFace, specifically ``google/ddpm-ema-church-256'', ``google/ddpm-cat-256'' and ``google/ddpm-ema-celebahq-256''. For the results reported in Table~\ref{tab:attackPDM}, we run 30 images for each victim model. Additionally, for generalizability in practical scenarios, we synthesize the data with half randomly selected from the originally trained dataset and another half from randomly crawled with keywords from the Internet.

\subsubsection{Baseline Methods and Evaluation Metrics.}
To the best of our knowledge, the previous methods have mainly focused on LDMs, and effective PDM attacks have not yet been developed, however, we still implement AdvDM ~\cite{liang2023adversarialexampledoesgood} with the proposed semantic loss by~\cite{salman2023raisingcostmaliciousaipowered, liang2023adversarialexampledoesgood, liang2023mistimprovedadversarialexamples, xue2024effectiveprotectiondiffusionbased} for comparison. Notably, Diff-Protect~\cite{xue2024effectiveprotectiondiffusionbased} proposed to minimize the semantic loss and is counterintuitively better than maximizing the semantic loss. We also adopt this method in attacking PDMs. To quantify the adversarial image visual quality, we adopt Structural Similarity (SSIM)~\cite{wang2004image}, Peak Signal-to-Noise Ratio (PSNR), and Learned Perceptual Image Patch Similarity (LPIPS) ~\cite{zhang2018unreasonable} as the evaluation metrics but negatively quantify the effectiveness of our attack. We also adopt the Image Alignment Score (IA)~\cite{kumari2023multi} that leverages CLIP~\cite{radford2021learning} to calculate the cosine similarity between two image encoder features. In distinguishing from the previous methods, to more faithfully reflect the attacking effectiveness, we fix the same seed of the random generator when generating clean and adversarial samples, then calculating the scores based on the paired samples.

\subsection{Attacking Effectiveness on PDMs}
\begin{figure*}[t]
\centering
\includegraphics[width=0.82\linewidth]{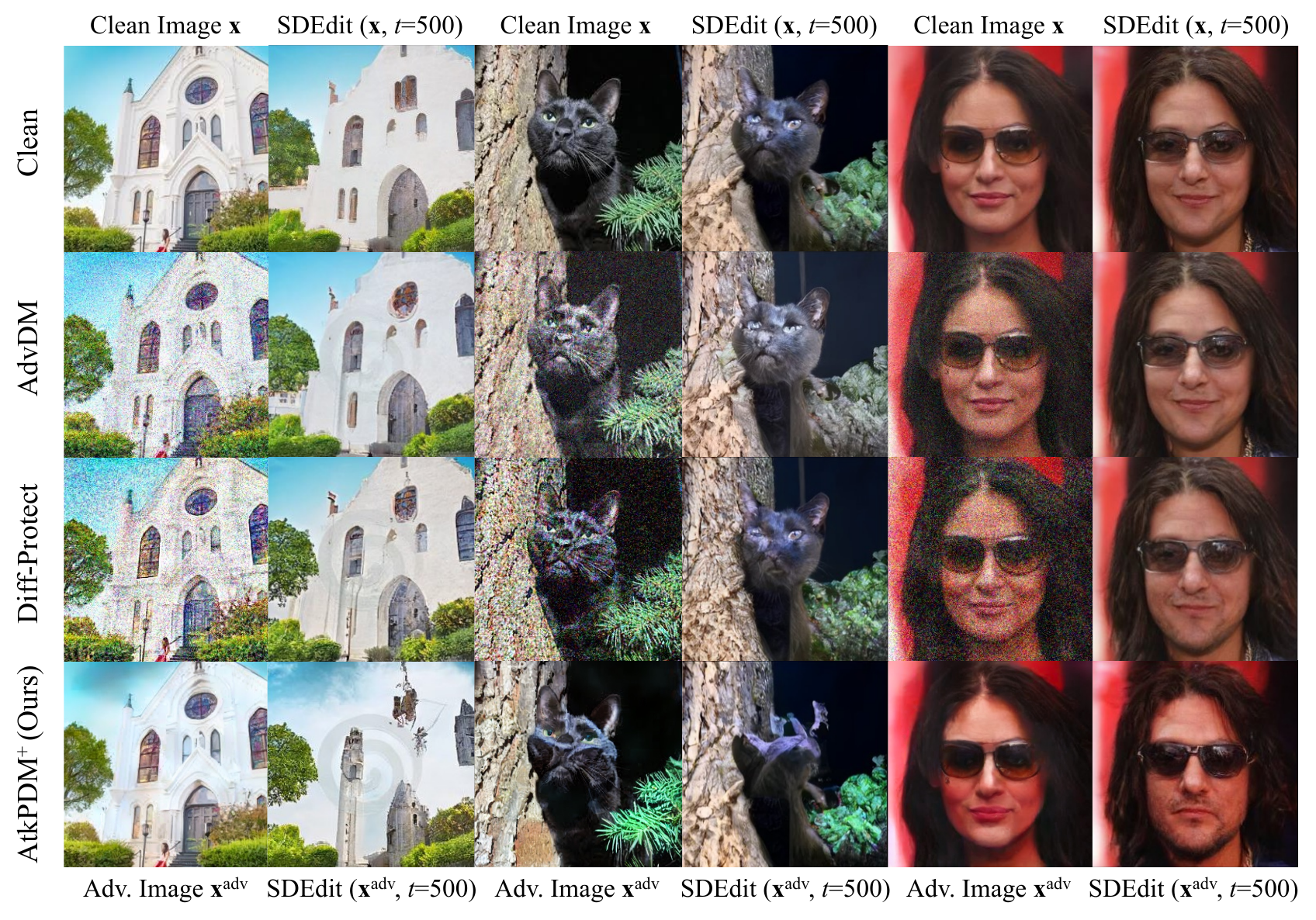}
\caption{Qualitative results compared to the previous methods. Our adversarial images can effectively corrupt the edited results without significant fidelity decrease. The same column shares the same random seed for fair comparisons.
}
\label{qualitative}
\end{figure*}

As quantitatively reported in Table~\ref{tab:attackPDM} and qualitative results in Figure~\ref{qualitative}, compared to the previous PGD-based methods incorporating semantic loss, i.e., negative training loss of diffusion models, our method exhibits superior performance in both adversarial image quality and attacking effectiveness. In addition, our reported numbers are generally stable, as reflected in lower standard deviation. It is worth noting that even if the adversarial image qualities of the PGD-based methods are far worse than ours, their attacking effectiveness still falls short, suggesting that PDMs are robust against traditional perturbation methods. This finding is also aligned with previous works~\cite{xue2024effectiveprotectiondiffusionbased,xue2024pixelbarrierdiffusionmodels}. For AtkPDM$^+$, combined with our latent optimization strategy, the adversarial image quality has been enhanced while slightly affecting the attacking effectiveness, still outperforming the previous methods. Besides unconditional PDMs, we also compare with the previous best method Diff-Protect against a conditional PDM DeepFloyd IF~\cite{DeepFloydIF}, reported in Table ~\ref{tab:deepfloydif}. Although the attacking effectiveness of Diff-Protect seems better than ours, this may be due to their adversarial image quality being severely corrupted during the attack. Hence, it cannot fulfill our two objectives simultaneously. In addition, our framework is extensible to attack LDMs, please refer to Appendix provided in the project page.

\begin{table}[t]
\footnotesize{
    \centering
    \begin{tabular}{lcccc}
        \toprule
        \multirow{2}{*}{Setting} & \multicolumn{4}{c}{Attacking Effectiveness} \\ 
         &  SSIM $\downarrow$ & PSNR $\downarrow$ & LPIPS $\uparrow$ & IA $\downarrow$ \\
        \midrule
        White Box & 0.79 & 30.05 & 0.33 & 0.81 \\
        
        Black Box & 0.86 & 30.25 & 0.29 & 0.85 \\
    
        \midrule
        Difference & 0.07 & 0.20 & 0.04 & 0.04 \\
        \bottomrule
    \end{tabular}
    \caption{Quantitative results of black box attack. We use the same set of adversarial images and feed to white box and black box models to examine the black box transferability.} 
\label{tab:blackBox}
}
\end{table}

\begin{table*}
    \centering
    \small{
    \begin{tabular}{lc|ccc|cccc}
        \toprule
        \multirow{2}{*}{Losses} & \multirow{2}{*}{VAE} & \multicolumn{3}{c|}{Adversarial Image Quality} & \multicolumn{4}{c}{Attacking Effectiveness} \\ 
         & & SSIM $\uparrow$ & PSNR $\uparrow$ & LPIPS $\downarrow$ & SSIM $\downarrow$ & PSNR $\downarrow$ & LPIPS $\uparrow$ & IA $\downarrow$ \\
        \midrule
        $\mathcal{L}_\text{semantic}$ &  & 0.37 $\pm$ 0.09  & 28.17 $\pm$ 0.22 & 0.73 $\pm$ 0.16  & 0.89 $\pm$ 0.05 & 31.06 $\pm$ 1.94 & 0.17 $\pm$ 0.09 & 0.93 $\pm$ 0.04 \\
        $\mathcal{L}_\text{semantic}$ & \checkmark & 0.80 $\pm$ 0.05 & \second{29.78} $\pm$ 0.42 & \second{0.17} $\pm$ 0.03 & 0.82 $\pm$ 0.05 & 30.43 $\pm$ 0.75 & 0.15 $\pm$ 0.06 & 0.92 $\pm$ 0.04 \\
        $\mathcal{L}_\text{semantic}$ + $\mathcal{L}_\text{fidelity}$ & \checkmark & \first{0.82} $\pm$ 0.05 & \first{30.30} $\pm$ 0.81 & \first{0.13} $\pm$ 0.03 & 0.90 $\pm$ 0.03 & 31.24 $\pm$ 1.19 & 0.08 $\pm$ 0.03 & 0.96 $\pm$ 0.02 \\
        $\mathcal{L}_\text{attack}$ + $\mathcal{L}_\text{fidelity}$ & & 0.75 $\pm$ 0.03 & 28.22 $\pm$ 0.10  & 0.26 $\pm$ 0.04 & \first{0.75} $\pm$ 0.04 & \first{29.61} $\pm$ 0.23 & \first{0.40} $\pm$ 0.05 & \first{0.76} $\pm$ 0.06 \\
        $\mathcal{L}_\text{attack}$ + $\mathcal{L}_\text{fidelity}$ & \checkmark & \second{0.81} $\pm$ 0.03 & 28.64 $\pm$ 0.19 & \first{0.13} $\pm$ 0.02 & \second{0.79} $\pm$ 0.04 & \second{30.05} $\pm$ 0.47 & \second{0.33} $\pm$ 0.07 & \second{0.81} $\pm$ 0.06 \\
        \bottomrule
    \end{tabular}
    }
    \caption{Quantitative results of ablation study. The best is marked in bold and the second best is underlined. Errors denote one standard deviation of all images in our test datasets.}
    \label{tab:loss_ablation}
\end{table*}

\subsection{Black Box Transferability}
We craft adversarial images with the proxy model, ``google/ddpm-ema-church-256'', in white-box settings and test their transferability against ``google/ddpm-bedroom-256'' model as black-box attacks.
Under identical validation settings, Table~\ref{tab:blackBox} reveals only a slight decrease in attack effectiveness metrics, suggesting black-box transferability.

\subsection{Robustness Against Defense Methods}
We examine the robustness of our approach against three widely recognized and effective adversarial defense methods. The quantitative results in Table~\ref{tab:defense} demonstrate that our method is robust against these three defense methods, with four metrics listed in Table~\ref{tab:defense} not worse than no defenses. Surprisingly, these defense methods even make the adversarial image more effective than cases without defense. We provide the implementation details of each defense method in the following sections.

\subsubsection{LDM Purification.}
Nie at. al. proposed DiffPure ~\cite{nie2022diffusionmodelsadversarialpurification} that leverages a pre-trained Diffusion Model to purify adversarial images targeting classifier models to defend effectively.
The purification process is essentially an unconditional SDEdit process with small forward $t$. Here, we use a pre-trained LDM (StableDiffusion v1.5) and $t=100$ to purify our adversarial image as a defense method.

\subsubsection{Crop and Resize.}
Noted by Diff-Protect, ``crop and resize'' is a simple yet the most effective defense method against their attacks on LDMs. We test our method against this defense using their settings, i.e., cropping 20\% of the adversarial image and resizing it to its original dimensions. 

\subsubsection{JPEG Compression.}
Sandoval-Segura et al.~\cite{sandoval2023jpeg} demonstrated that JPEG compression is a simple yet effective adversarial defense method. In our experiments, we implement the JPEG compression at a quality setting of 25\%.

\subsection{Effectiveness of Latent Optimization via VAE}
We first incorporate our VAE latent optimization strategy in the previous semantic-loss-based methods. 
From Table~\ref{tab:loss_ablation}, without using $\mathcal{L}_\text{fidelity}$, latent optimization has significantly enhanced the adversarial image quality and even slightly improved the attacking effectiveness. Adopting latent optimization in our approach enhances visual quality with a negligible decrease in attacking effectiveness. Surprisingly, incorporating our $\mathcal{L}_\text{fidelity}$ with current PGD-based method will drastically decrease the adversarial image quality despite its attack performing better than ours. This may be due to different constrained optimization problem settings.

\section{Conclusion}
This paper presents the first framework to protect against image manipulation by Pixel-domain Diffusion Models (PDMs). While denoising UNets withstand traditional PGD attacks, their feature space remains vulnerable. Our feature attacking loss exploits these vulnerabilities, generating adversarial images that mislead PDMs, resulting in corrupted output. We approach this image protection problem as a constrained optimization problem, solving it through alternating optimization. Furthermore, our latent optimization strategy via VAE enhances the naturalness of our adversarial images. Extensive experiments validate the efficacy of our method, achieving state-of-the-art performance in attacking PDMs. 

\section*{Acknowledgements}
This research is supported by National Science and Technology Council, Taiwan (R.O.C) under the grant numbers NSTC-113-2634-F-002-007, NSTC-112-2222-E-001-001-MY2, NSTC-113-2634-F-001-002-MBK, NSTC-113-2221-E-002-201, and Academia Sinica under the grant number of AS-CDA-110-M09. We thank to National Center for High-performance Computing (NCHC) of National Applied Research Laboratories (NARLabs) in Taiwan for providing computational and storage resources.

\bibliography{aaai25}

\clearpage
\appendix
\begin{center}
    \LARGE{\textbf{Appendix}}
\end{center}

\section{More Implementation Details}
The feature extractor for calculating $\mathcal{L}_\text{fidelity}$ is VGG16~\cite{Simonyan2014VeryDC} with IMAGENET1K-V1 checkpoint. We use the SDEdit with the forward step $t=500$ for our main study results as it balances faithfulness to the original image and flexibility for editing. Empirically, we choose to randomly sample the forward step $t \sim [0, 500]$ to enhance the optimization efficiency. The average time to optimize 300 steps for an image on a single Nvidia Tesla V100 is about 300 seconds. The estimated average memory usage is about 24GB. Table \ref{tab:step_size} provides the details of the step sizes that we use to attack different models. 

\begin{table}[h]
\footnotesize{
    \centering
    \begin{tabular}{lccc}
        \toprule
        \multirow{2}{*}{Models} & \multicolumn{2}{c}{Step Size} \\ 
         &  $\gamma_\text{attack}$ & $\gamma_\text{fidelity}$ \\
        \midrule
        google/ddpm-ema-church-256 & $100/255$ & $40/255$ \\
        google/ddpm-cat-256 & $100/255$ & $5/255$ \\
        google/ddpm-ema-celebahq-256 & $50/255$ & $35/255$ \\  
        
        \bottomrule
    \end{tabular}
    \caption{The step sizes used for different models during optimization.} 
\label{tab:step_size}
}
\end{table}

\section{More Experimental Results}
\subsection{Attack Effectiveness on Latent Diffusion Models}
We propose the feature representation attacking loss which can be adapted to target any UNet-based diffusion models. Hence, it is applicable to attack LDM using our proposed framework. We follow the evaluation settings of the previous work~\cite{xue2024effectiveprotectiondiffusionbased} for fair comparisons. Quantitative results are shown in Table~\ref{tab:attackLDM}. Compared to previous LDM-specified methods~\cite{liang2023adversarialexampledoesgood, liang2023mistimprovedadversarialexamples, xue2024effectiveprotectiondiffusionbased}, our method could achieve comparable results. This finding reflects the general vulnerability in UNet-based diffusion models that can be exploited to craft adversarial images against either PDMs or LDMs. 

\begin{table*}[t]
    \centering
    \small{
    \begin{tabular}{ll|ccc|cccc}
        \toprule
        & \multirow{2}{*}{Methods} & \multicolumn{3}{c|}{Adversarial Image Quality} & \multicolumn{4}{c}{Attacking Effectiveness} \\ 
         & & SSIM $\uparrow$ & PSNR $\uparrow$ & LPIPS $\downarrow$ & SSIM $\downarrow$ & PSNR $\downarrow$ & LPIPS $\uparrow$ & IA $\downarrow$ \\
        \midrule
        \multirow{6}{*}{\rotatebox{90}{Church}}
        & AdvDM
        & \first{0.85} $\pm$ 0.03 & \first{30.42} $\pm$ 0.15 & \second{0.23} $\pm$ 0.06 & 0.19 $\pm$ 0.05 & 28.00 $\pm$ 0.16 & 0.71 $\pm$ 0.04 & 0.49 $\pm$ 0.06 \\
        & Mist
        & 0.81 $\pm$  0.03 & 29.45 $\pm$ 0.13 & 0.25 $\pm$ 0.05 & \first{0.14} $\pm$ 0.03 & \first{27.95} $\pm$ 0.13 & \first{0.76} $\pm$ 0.04 & \second{0.48} $\pm$ 0.05 \\
        & Diff-Protect
        & 0.79 $\pm$ 0.03 & 29.92 $\pm$ 0.15 & 0.24 $\pm$ 0.06 & \second{0.15} $\pm$ 0.03 & 28.00 $\pm$ 0.14 & 0.71 $\pm$ 0.04 & \second{0.48} $\pm$ 0.05 \\
        & AtkPDM (Ours) & \second{0.82} $\pm$ 0.02 & \second{30.40} $\pm$ 0.27 & 0.24 $\pm$ 0.05 & \first{0.14} $\pm$ 0.03 & \second{27.96} $\pm$ 0.17 & \second{0.74} $\pm$ 0.02 & \first{0.47} $\pm$ 0.04  \\
        & AtkPDM$^+$ (Ours) & 0.61 $\pm$ 0.07 & 29.17 $\pm$ 0.32 & \first{0.20} $\pm$ 0.02 & 0.27 $\pm$ 0.06 & 28.07 $\pm$ 0.18 & 0.66 $\pm$ 0.05 & 0.51 $\pm$ 0.06 \\
        \midrule
        \multirow{6}{*}{\rotatebox{90}{Cat}}
        & AdvDM
        & \first{0.86} $\pm$ 0.04 & \second{30.68} $\pm$ 0.24 & \second{0.25} $\pm$ 0.09 & 0.21 $\pm$ 0.05 & 28.03 $\pm$ 0.21 & 0.70 $\pm$ 0.07 & \second{0.53} $\pm$ 0.04 \\
        & Mist
        & 0.81 $\pm$ 0.04 & 29.63 $\pm$ 0.22 & 0.27 $\pm$ 0.08 & \first{0.14} $\pm$ 0.04 & \first{27.96} $\pm$ 0.17 & \first{0.77} $\pm$ 0.06 & \first{0.52} $\pm$ 0.04 \\
        &
        Diff-Protect
        & 0.78 $\pm$ 0.05 & 30.12 $\pm$ 0.24 & 0.27 $\pm$ 0.08 & \second{0.16} $\pm$ 0.05 & \first{27.96} $\pm$ 0.15 & \second{0.72} $\pm$ 0.06 & \first{0.52} $\pm$ 0.03 \\
        & AtkPDM (Ours) & \second{0.84} $\pm$ 0.02 & \first{30.79} $\pm$ 0.49 & \second{0.25} $\pm$ 0.07 & 0.18 $\pm$ 0.04 & 28.00 $\pm$ 0.19 & \second{0.72} $\pm$ 0.05 & \first{0.52} $\pm$ 0.03  \\
        & AtkPDM$^+$ (Ours) & 0.68 $\pm$ 0.13 & 29.68 $\pm$ 0.74 & \first{0.16} $\pm$ 0.03 & 0.31 $\pm$ 0.10 & 28.13 $\pm$ 0.27 & 0.64 $\pm$ 0.06 & 0.54 $\pm$ 0.04 \\
        \midrule
        \multirow{6}{*}{\rotatebox{90}{Face}}
        & AdvDM
        & \first{0.83} $\pm$ 0.02 & \second{30.81} $\pm$ 0.22 & 0.32 $\pm$ 0.06 & 0.26 $\pm$ 0.05 & 28.07 $\pm$ 0.28 & 0.74 $\pm$ 0.05 & 0.47 $\pm$ 0.07 \\
        & Mist
        & 0.79 $\pm$ 0.03 & 29.75 $\pm$ 0.22 & 0.34 $\pm$ 0.06 & \first{0.19} $\pm$ 0.05 & \first{27.99} $\pm$ 0.21 & \first{0.81} $\pm$ 0.05 & 0.46 $\pm$ 0.08 \\
        &
        Diff-Protect
        & 0.74 $\pm$ 0.04 & 30.34 $\pm$ 0.13 & 0.33 $\pm$ 0.06 & \second{0.21} $\pm$ 0.05 & \second{28.03} $\pm$ 0.21 & 0.76 $\pm$ 0.06 & \second{0.45} $\pm$ 0.07 \\
        & AtkPDM (Ours) & \first{0.83} $\pm$ 0.02 & \first{31.21} $\pm$ 0.44 & \second{0.31} $\pm$ 0.05 & \second{0.21} $\pm$ 0.04 & \second{28.03} $\pm$ 0.26 & \second{0.78} $\pm$ 0.04 & \first{0.44} $\pm$ 0.06  \\
        & AtkPDM$^+$ (Ours) & \second{0.82} $\pm$ 0.05 & 30.05 $\pm$ 0.51 & \first{0.14} $\pm$ 0.03 & 0.41 $\pm$ 0.08 & 28.24 $\pm$ 0.39 & 0.63 $\pm$ 0.07 & 0.52 $\pm$ 0.07 \\
        \bottomrule
    \end{tabular}
    }
    \caption{Quantitative results in attacking LDM. The best is marked in bold and the second best is underlined. Errors denote one standard deviation of all images in our test datasets.}
    \label{tab:attackLDM}
\end{table*}

\begin{figure}[t]
\centering
\includegraphics[width=1\linewidth]{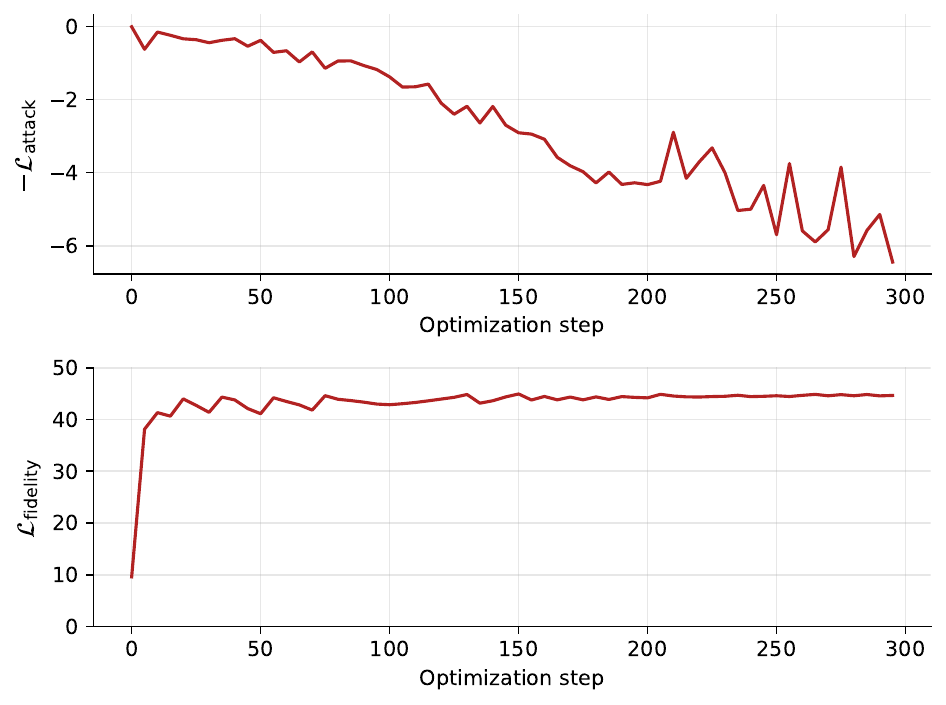}
\caption{Loss curves of our $\mathcal{L}_\text{attack}$ and $\mathcal{L}_\text{fidelity}$ against optimization step.}
\label{supp:loss_curve}
\end{figure}

\subsection{Qualitative Demonstration of Corrupting UNet Feature during Sampling}
\begin{figure*}[h]
\centering
\includegraphics[width=0.9\linewidth]{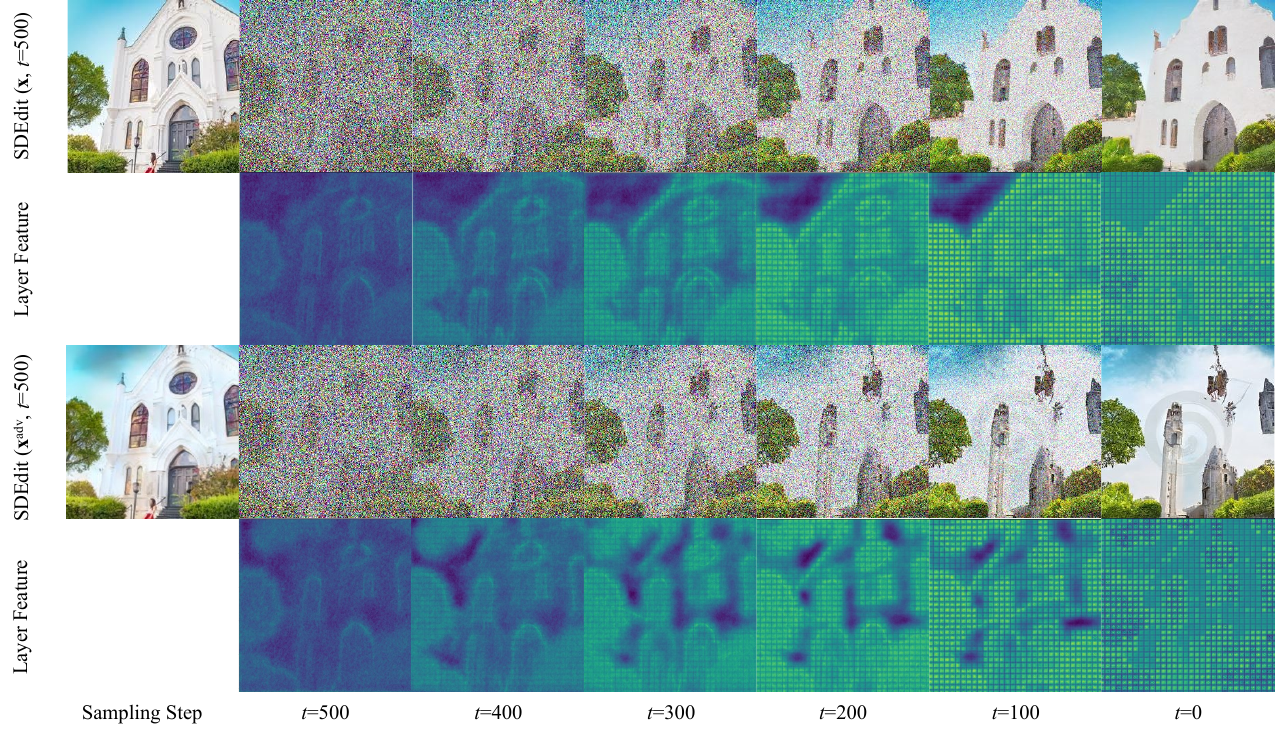}
\caption{Qualitative example of corrupting feature representations in UNet: as the denoising process proceeds, the similarity of the feature map decreases, suggesting the representation is corrupted.}
\label{fig:feature_visulization}
\end{figure*}

We qualitatively show an example of our attack effectiveness regarding UNet representation discrepancies in Figure~\ref{fig:feature_visulization}. We compare a clean and an adversarial image using the same denoising process. Then, we take the feature maps of the second-last decoder block layer, close to the final predicted noise, to demonstrate their recognition of input image semantics.
The results in Figure~\ref{fig:feature_visulization} show that from $t$ = 500, the feature maps of each pair start with a similar structure, then as the $t$ decreases, the feature maps gradually have higher discrepancies, suggesting our method, by attacking the middle representation of UNet, can effectively disrupt the reverse denoising process and mislead to corrupted samples.

\subsection{Qualitative Results of Loss Ablation}
\begin{figure*}[h]
\centering
\includegraphics[width=1\linewidth]{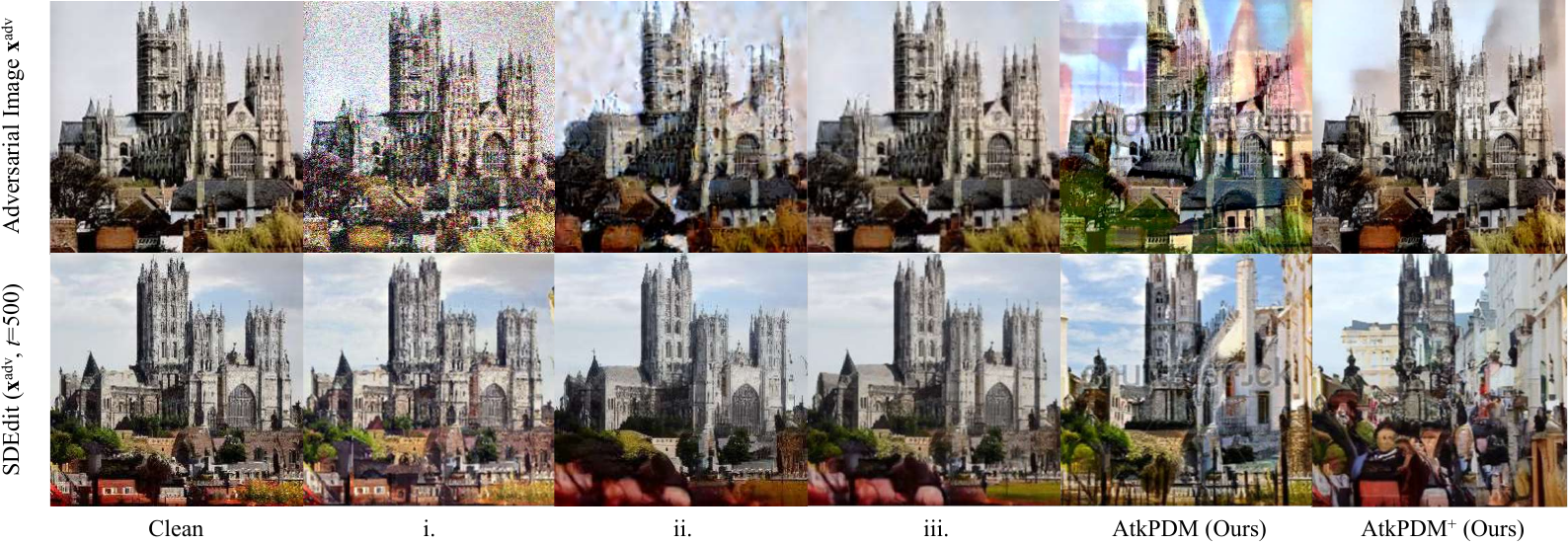}
\caption{Qualitative example of different loss configurations. i. only semantic loss; ii. semantic loss and latent optimization; iii. semantic loss, $\mathcal{L}_\text{fidelity}$ and latent optimization.}
\label{fig:loss_ablation}
\end{figure*}

Figure~\ref{fig:loss_ablation} presents qualitative results of loss ablation where i., ii., and iii. indicate performing PGAscent with different configurations. i. utilizes only semantic loss; ii. utilizes semantic loss with our latent optimization strategy; iii. utilizes both semantic loss, our proposed $\mathcal{L}_\text{fidelity}$ and latent optimization. The results show that our $\mathcal{L}_\text{fidelity}$ and latent optimization can enhance the adversarial image quality of PGAscent. Moreover, comparing our proposed two methods, AtkPDM$^+$ generates a more natural adversarial image than AtkPDM while maintaining attack effectiveness.

\subsection{Example of Loss Curves}
Figure \ref{supp:loss_curve} shows an example of our loss trends among optimization steps. $\mathcal{L}_\text{attack}$ has decreasing trend as the optimization step increases. $\mathcal{L}_\text{fidelity}$ has an increasing trend and converges to satisfy the constraint of the attack budget $\delta$.

\subsection{Different Forward Time-step Sampling}
\begin{figure*}[h]
\centering
\includegraphics[width=0.9\linewidth]{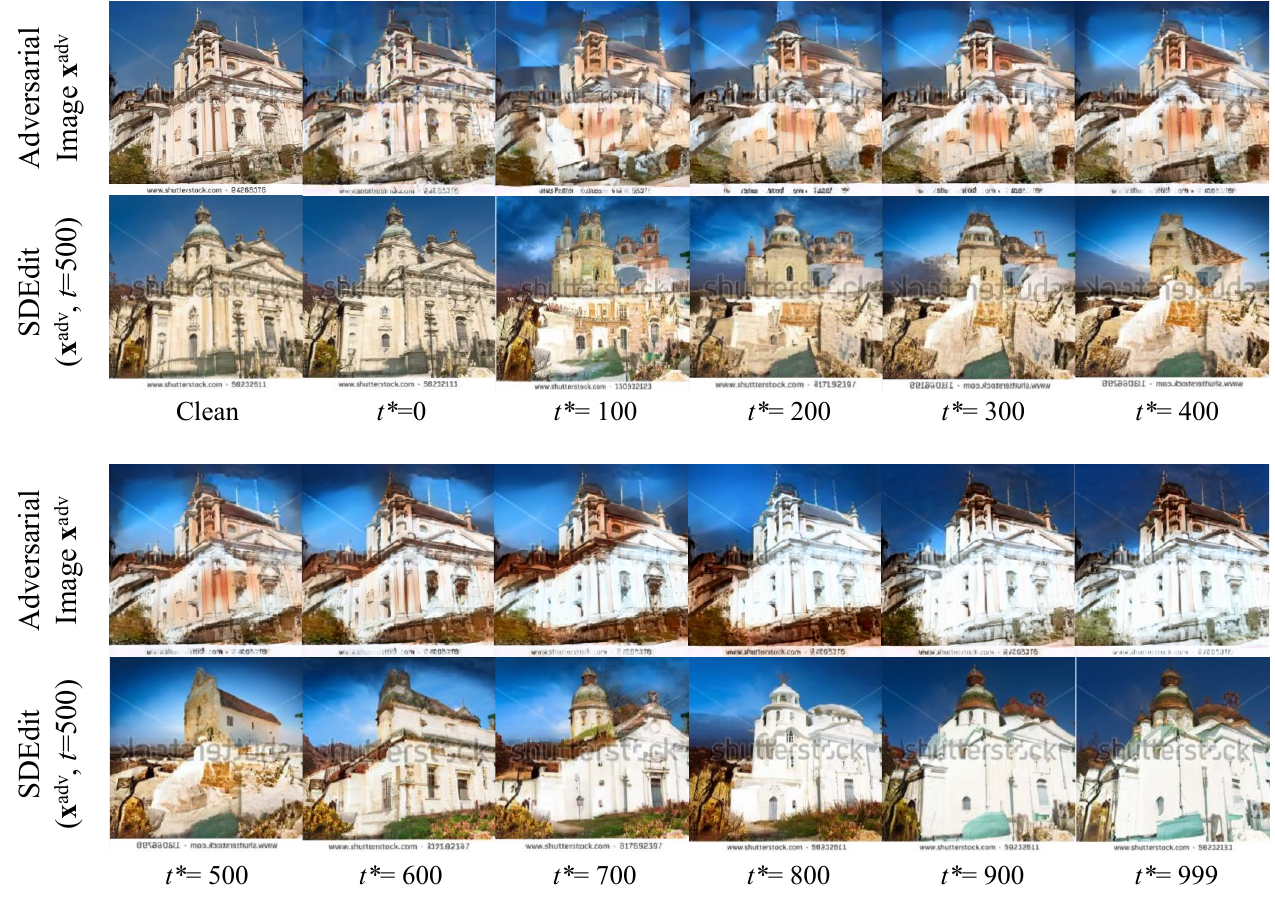}
\caption{Qualitative results of optimizing different fixed diffusion forward steps $t^*$.}
\label{fig:different_timestep}
\end{figure*}

When using Monte Carlo sampling for optimization, the forward time step $t^*$ is sampled uniformly. We study the scenario that when $t^*$ is fixed for optimization. As shown in Figure~\ref{fig:different_timestep}, a primary result shows that when attacking $t^*=400$ to $t^*=500$, the attacking effectiveness is better than other time steps. In practice, we can not know user-specified $t^*$ for editing in advance; however, this suggests that diffusion models have a potential temporal vulnerability that can be further exploited to increase efficiency.

\subsection{More Qualitative Results}
\begin{figure*}
\centering
\begin{subfigure}{1\linewidth}
    \centering
    \includegraphics[width=0.9\linewidth]{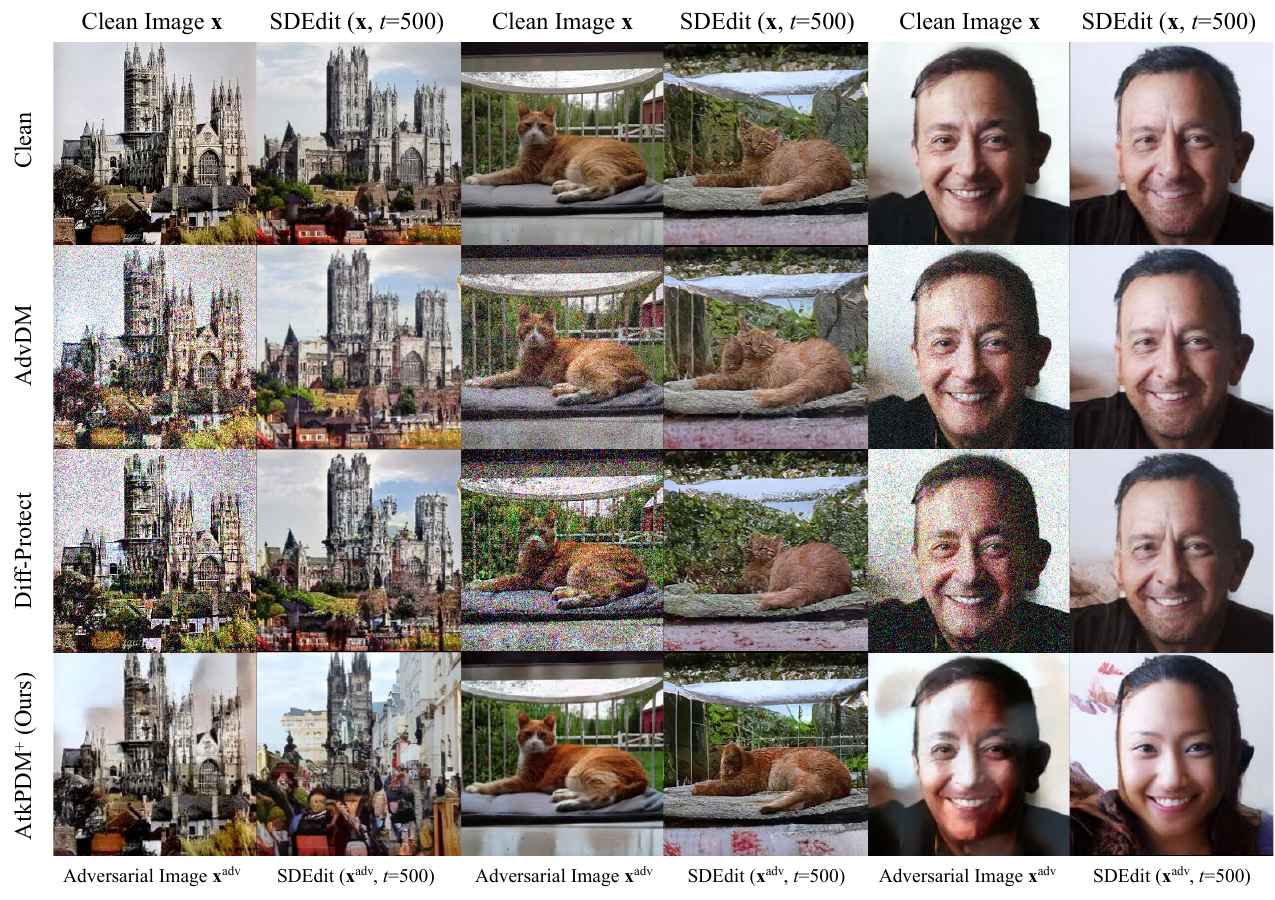}
    \label{fig:qualitative_results_1}
\end{subfigure}

\begin{subfigure}{0.9\linewidth}
    \centering
    \includegraphics[width=1\linewidth]{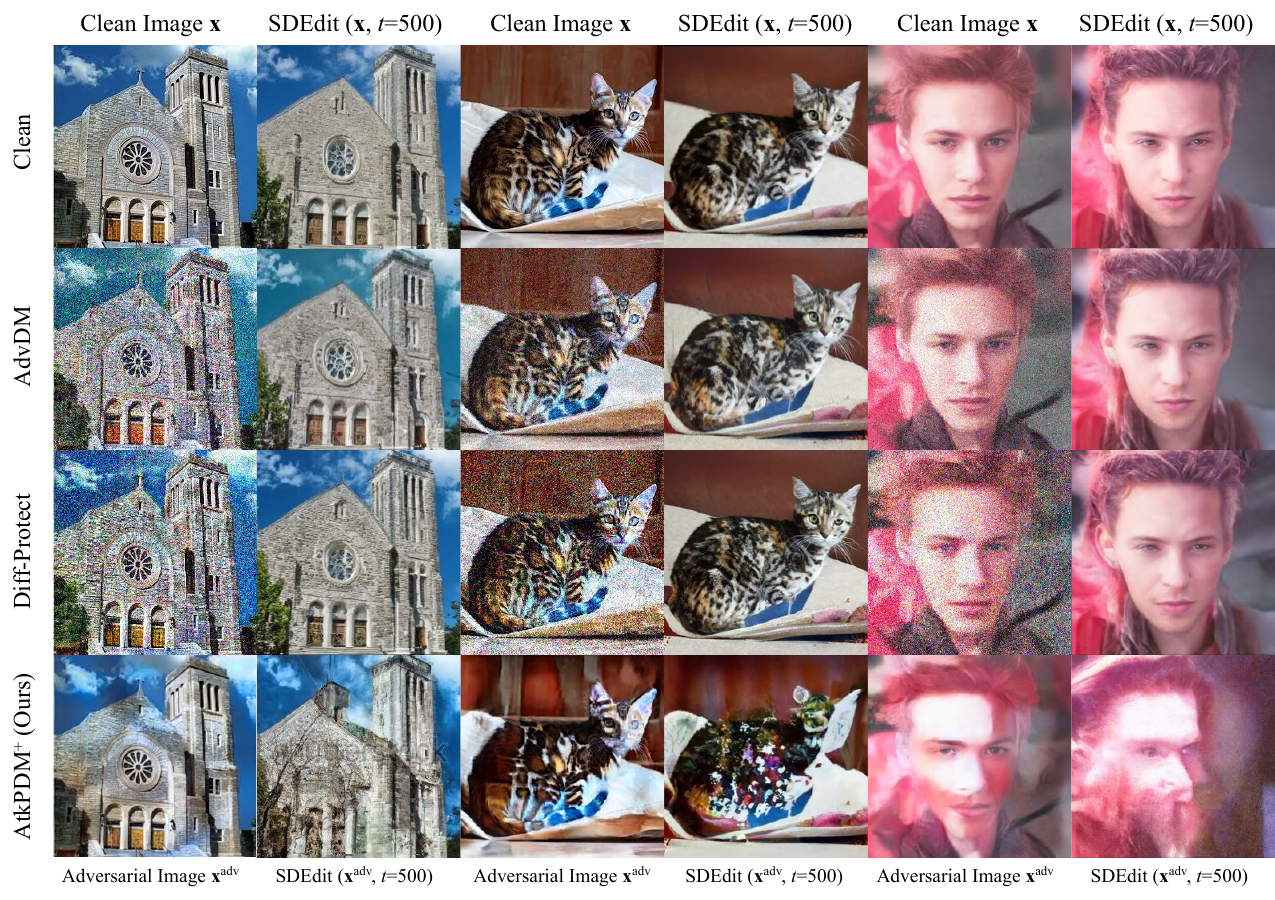}
    \label{fig:qualitative_results_2}
\end{subfigure}
\caption{Qualitative results compared to previous methods: our adversarial images can effectively corrupt the edited results without significant fidelity decrease. The same column shares the same random seed for fair comparison.}
\label{supp:qualitative}
\end{figure*}

We provide more qualitative results in Figure~\ref{supp:qualitative} to showcase that our method can significantly change or corrupt the generated results with little modification on adversarial images. In contrast, previous methods add obvious perturbation to adversarial images but still fail to change the edited results to achieve the safeguarding goal.

\section{Backgrounds of Diffusion Models}
Score-based models and diffusion models allowing generate samples starting from easy-to-sample Gaussian noise to complex target distributions. Starting from Gaussian noise, the sampling process iteratively applies the score function, i.e., $\nabla_{\mathbf{x}} \log p(\mathbf{x})$ of the complex target distribution $p(\mathbf{x})$ to generate the sample from $p(\mathbf{x})$. The exact estimation of the ground truth score function is intractable since the score function is the derivative of the target distribution $p(\mathbf{x})$. However, we can approximate the score function without directly dealing with $p(\mathbf{x})$. Song et al. proposed score-based models~\cite{song2020sliced} to learn the score function effectively via score matching. Ho et al. proposed Denoising Diffusion Probability Model (DDPM)~\cite{ho2020denoising}, providing another perspective on learning score function with noise perturbed data, allowing more effective low-density area estimation and improving the mode diversity, thereby capable of generating highly sophisticated data, e.g., natural images. In a nutshell, training DDPM involves perturbing data with Gaussian noise in different timestep-controlled variance schedules, i.e., forward diffusion, and a parametrized model $\epsilon_{\theta}(\mathbf{x}_t, t)$ will learn to predict the added noise conditioning on noisy data $\mathbf{x}_t$ and current noise level  $t$. Sampling with learned DDPM starts with random noise and iteratively applies the model $\epsilon_{\theta}(\mathbf{x}_t, t)$ to denoise, i.e., reverse diffusion sampling, thereby generating a sample from the learned distribution. Specifically, for forward diffusion, we perturb the data with a linear combination of Gaussian noise and clean data as $\mathbf{x}_t = \sqrt{\bar{\alpha}_t} \mathbf{x} + \sqrt{1 - \bar{\alpha}_t} \epsilon_t$ via a scheduler $\bar{\alpha}_t$ controlling the strength of added noise, here $t \in[0, T]$ and $\epsilon_t \sim \mathcal{N}(\mathbf{0}, \mathbf{I})$, note that when $t$ reaches $T$, the perturbed data $\mathbf{x}_t$ become Gaussian noise. The training objective of the $\epsilon_{\theta}$ is defined as the noise prediction MSE $\mathbb{E}_{t, \mathbf{x}, \epsilon_t}[\| \epsilon_t - \epsilon_{\theta}(\mathbf{x}_t, t) \|_2^2]$. In sampling with diffusion models, Song et.al proposed DDIM~\cite{song2021denoisingdiffusionimplicitmodels} that generalized the DDPM sampling formulation as:
\begin{align*}
    \mathbf{x}_{t-1} &= \sqrt{\bar{\alpha}_{t - 1}} \left(\frac {\mathbf{x}_t - \sqrt{1- \bar{\alpha}_t}\epsilon_{\theta}(\mathbf{x}_t, t)}{\sqrt{\bar{\alpha}_t}} \right) \\ & \quad + \sqrt{1-\bar{\alpha}_{t-1}-\sigma^{2}_t}\epsilon_{\theta}(\mathbf{x}_t, t) + \sigma_t \epsilon_t.
\end{align*}

The first term of the right-hand side of the equation represents direct clean sample estimation $\hat{\mathbf{x}}_0$ from noisy sample $\mathbf{x}_t$ which is derived from Tweedie’s formula\cite{efron2011tweedie}. Therefore, the noise prediction can bridge with the score function via Tweedie’s formula, where the denoising objective and score-matching objective are identical.

\section{Details of Our Proposed Algorithm}

\subsection{2-Wasserstein Distance Between Two Normal Distribution}
Consider the normal distributions $\mathcal{N}_t:=\mathcal{N}(\mu_t, \Sigma_t)$ and $\mathcal{N}_t^{\adv}:=\mathcal{N}(\mu_t^{\adv}, \Sigma_t^{\adv})$. Let $\Pi(\mathcal{N}_t, \mathcal{N}_t^{\adv})$ denote a joint distribution over the product space $\mathbb{R}^n \times \mathbb{R}^n$. The 2-Wasserstein distance between $\mathcal{N}_t$ and $\mathcal{N}_t^{\adv}$ is defined as:
\begin{align*}
    \mathcal{W}_2^2(\mathcal{N}_t, \mathcal{N}_t^{\adv})
    =\min_{\pi \in \Pi(\mathcal{N}_t, \mathcal{N}_t^{\adv})} \int \|\mathbf{f}_t-\mathbf{f}_t^{\adv}\|_2^2 \text{d}\pi(\mathbf{f}_t, \mathbf{f}_t^{\adv}).
\end{align*}

Using properties of the mean and covariance, we have the following identities:
\begin{align*}
    &\int \|\mu_t-\mu_t^{\adv}\|_2^2 \text{d}\pi(\mathbf{f}_t, \mathbf{f}_t^{\adv}) = \|\mu_t-\mu_t^{\adv}\|_2^2, \\
    &\int \|\mathbf{f}_t-\mu_t\|_2^2 \text{d}\pi(\mathbf{f}_t, \mathbf{f}_t^{\adv}) = \trac(\Sigma_t), \\
    &\int \|\mathbf{f}_t^{\adv}-\mu_t^{\adv}\|_2^2 \text{d}\pi(\mathbf{f}_t, \mathbf{f}_t^{\adv}) = \trac(\Sigma_t^{\adv}), \\
    &\int (\mathbf{f}_t-\mu_t)^\top(\mathbf{f}_t^{\adv}-\mu_t^{\adv}) \text{d}\pi(\mathbf{f}_t, \mathbf{f}_t^{\adv}) \\
    &=\trac\left(\mathbb{E}[(\mathbf{f}_t-\mu_t)(\mathbf{f}_t^{\adv}-\mu_t^{\adv})^\top\right).
\end{align*}
Thus, the 2-Wasserstein distance can be expressed as:
\begin{equation*}
\begin{aligned}
    \mathcal{W}_2^2(\mathcal{N}_t, \mathcal{N}_t^{\adv})
    &=\|\mu_t-\mu_t^{\adv}\|_2^2 \\
    + \trac{(\Sigma_t)} &+ \trac{(\Sigma_t^{\adv})}
    -2\max_{J \succeq 0} \trac(C),
\end{aligned}
\end{equation*}
where $J$ is the joint covariance matrix of $\mathcal{N}_t$ and $\mathcal{N}_t^{\adv}$, defined as:
\begin{align*}
J = \left[
\begin{array}{cc}
    \Sigma_t & C \\
    C^\top   & \Sigma_t^{\adv}
\end{array}
\right],
\end{align*}
and $C$ is the covariance matrix between $\mathcal{N}_t$ and $\mathcal{N}_t^{\adv}$:
\begin{align*}
C = \mathbb{E}
\left[
(\mathbf{f}_t-\mu_t)(\mathbf{f}_t^{\adv}-\mu_t^{\adv})^{\top}
\right].
\end{align*}

By the Schur complement, the problem can be formulated as a semi-definite programming (SDP) problem:
\begin{equation*}
\begin{aligned}
&\text{maximum} \quad \trac(C), \\
&\text{subject to } \quad
\Sigma_t - C^\top (\Sigma_t^{\adv})^{-1} C \succeq 0.
\end{aligned}
\end{equation*}
The closed-form solution for $C$ derived from the SDP is:
\begin{equation*}
    C=
    \Sigma_t^{\frac{1}{2}}
    (\Sigma_t^{\frac{1}{2}}\Sigma_t^{\adv}\Sigma_t^{\frac{1}{2}})^\frac{1}{2}
    \Sigma_t^{-\frac{1}{2}}.
\end{equation*}

Finally, the closed-form solution for the 2-Wasserstein distance between the two normal distributions is given by:
\begin{equation}
\begin{aligned}
    \mathcal{W}_2^2(\mathcal{N}_t, \mathcal{N}_t^{\adv})
    &=\|\mu_t-\mu_t^{\adv}\|_2^2 \\
    + \trac{(\Sigma_t)} &+ \trac{(\Sigma_t^{\adv})}
    -2(\Sigma_t^{\frac{1}{2}}\Sigma_t^{\adv}\Sigma_t^{\frac{1}{2}})^\frac{1}{2}.
\end{aligned}
\end{equation}

\subsection{Alternating Optimization}
Let $\mathbf{y} = \mathbf{x}^{\adv}$, by Lagrange relaxation \cite{liu2023instruct2attack}, the objective function can be expressed as:    
\begin{equation*}
F(\mathbf{x}, \mathbf{y}) = F_\text{attack}(\mathbf{x}, \mathbf{y}) + \lambda F_\text{fidelity}(\mathbf{x}, \mathbf{y}),
\end{equation*}
where $\lambda > 0$ is the Lagrange multiplier and $F_\text{attack}$, $F_\text{fidelity}$ are defined as
\begin{align*}
    F_\text{attack}(\mathbf{x}, \mathbf{y})
    &=-\mathcal{L}_\text{attack}(
    \mathcal{F}(\mathbf{x}, t, \epsilon), \mathcal{F}(\mathbf{y}, t, \epsilon^{\adv})), \\
    F_\text{fidelity}(\mathbf{x}, \mathbf{y})
    &=\max (\epsilon-\mathcal{L}_\text{fidelity}(\mathbf{x}, \mathbf{y}), \mathbf{0}).
\end{align*}

The optimization is carried out in an alternating manner as follows:
\begin{align}
    & \mathbf{y}^{i+\frac{1}{2}} = \argmin_{\mathbf{y}} \left(F_\text{attack}(\mathbf{x}, \mathbf{y}) + \lambda F_\text{fidelity}(\mathbf{x}, \mathbf{y}^{i})\right), \label{eq:F1} \\
    & \mathbf{y}^{i+1} = \argmin_{\mathbf{y}} \left(F_\text{attack}(\mathbf{x}, \mathbf{y}^{i+\frac{1}{2}}) + \lambda F_\text{fidelity}(\mathbf{x}, \mathbf{y})\right). \label{eq:F2}
\end{align}

To solve Equation \ref{eq:F1}, we employ the Fast Gradient Sign Method (FGSM) \cite{ian2015fgsm}. The update is given by:
\begin{equation*}
\mathbf{y}^{i+1/2} = \mathbf{y}^{i} - \gamma_\text{attack} \sign \left(\nabla_{\mathbf{y}^i} F_\text{attack}(\mathbf{x}, \mathbf{y}^i)\right).
\end{equation*}

For Equation \ref{eq:F2}, we utilize Gradient Descent, resulting in the following update:
\begin{align*}
\mathbf{y}^{i+1} = \mathbf{y}^{i+\frac{1}{2}} - \Tilde{\gamma}_\text{fidelity} \nabla_{\mathbf{y}^{i+\frac{1}{2}}} \lambda F_\text{fidelity}(\mathbf{x}, \mathbf{y}^{i+\frac{1}{2}}) \\
= \mathbf{y}^{i+\frac{1}{2}} - \gamma_\text{fidelity} \nabla_{\mathbf{y}^{i+\frac{1}{2}}} F_\text{fidelity}(\mathbf{x}, \mathbf{y}^{i+\frac{1}{2}}).
\end{align*}
Note that the gradient of $F_\text{fidelity}$ can be derived as follows:
\begin{align*}
    \nabla_{\mathbf{y}} F_\text{fidelity}(\mathbf{x}, \mathbf{y})
    = \mathbb{I}_\mathcal{C'} \cdot
    \nabla_{\mathbf{x}^{\adv}_t} \mathcal{L}_\text{fidelity}(\mathbf{x}, \mathbf{y}),
\end{align*}
where $\mathbb{I}_\mathcal{C'}$ is indicator function with constraint $\mathcal{C} = {\{\mathbf{y} \in \mathcal{M} \mid \mathcal{L}_\text{fidelity}(\mathbf{x}, \mathbf{y}) \leq \epsilon\}}$.\\

Please note that after references, we also provide more results presented in Figures~\ref{fig:loss_ablation}, ~\ref{fig:different_timestep}, ~\ref{supp:qualitative}, and~\ref{supp:loss_curve}, please refer to subsequent pages.

\subsection{AtkPDM Algorithm without Latent Optimization}

\begin{algorithm}[H]
    \caption{AtkPDM}
    \label{alg:attdpm}
    \small{
    \begin{algorithmic}[1] 
        \STATE{\textbf{Input:}
        Image to be protected $\mathbf{x}$, attack budget $\delta > 0$, and step size $\gamma_\text{attack}, \gamma_\text{fidelity}>0$}
        \STATE{\textbf{Initialization:} $\mathbf{x}^{\adv} \leftarrow \mathbf{x}$, $L_\text{attack} \leftarrow \infty$}
        \WHILE{$L_\text{attack}$ not convergent}
            \STATE{Sample timestep: $t \sim [0, T]$}
            \STATE{Sample noise: $\epsilon, \epsilon^{\adv} \sim \normaldist$}
            \STATE{Compute original noisy sample: \\
            $\mathbf{x}_t \leftarrow \mathcal{F}(\mathbf{x}, t, \epsilon)$}
            \STATE{Compute adversarial noisy sample: \\
            $\mathbf{x}^{\adv}_t \leftarrow \mathcal{F}(\mathbf{x}^{\adv}, t, \epsilon^{\adv})$}
            \STATE{Update $\mathbf{x}^{\adv}$ by Gradient Descent: \\
            $\mathbf{x}^{\adv} \leftarrow \mathbf{x}^{\adv} -
            \gamma_\text{attack} \sign(\nabla_{\mathbf{x}^{\adv}} (-\mathcal{L}_\text{attack}(\mathbf{x}^{\adv}_t, {\mathbf{x}_t})))$}            \WHILE{$\mathcal{L}_\text{fidelity}(\mathbf{x}^{\adv}, \mathbf{x}) > \delta$}
            \STATE{$\mathbf{x}^{\adv} \leftarrow \mathbf{x}^{\adv} -
            \gamma_\text{fidelity} \nabla_{\mathbf{x}^{\adv}} \mathcal{L}_\text{fidelity}(\mathbf{x}^{\adv}, \mathbf{x})$}
            \ENDWHILE
        \ENDWHILE
        \RETURN {$\mathbf{x}^{\adv}$}
    \end{algorithmic}
    }
\end{algorithm}

\section{Limitations}
While our method can deliver acceptable attacks on PDMs, its visual quality is still not directly comparable to the results achieved on LDMs, indicating room for further improvement. More generalized PDM attacks should be further explored. 

\section{Societal Impacts}
Our work will not raise potential concerns about diffusion model abuses. Our work is dedicated to addressing these issues by safeguarding images from being infringed.

\end{document}